\documentclass[11pt,a4paper]{article}
\usepackage[hyperref]{emnlp-ijcnlp-2019}
\usepackage{times}
\usepackage{latexsym}
\usepackage{kky}
\usepackage{url}
\usepackage{graphicx}
\usepackage{dblfloatfix}
\usepackage{amssymb}
\usepackage{multirow}
\usepackage{amsmath}
\usepackage{tabulary}
\usepackage{booktabs}
\usepackage{makecell}

\aclfinalcopy 

\newcommand{\metric}{REO}

\title{REO-Relevance, Extraness, Omission: A Fine-grained Evaluation for Image Captioning}

\author{Ming Jiang$^{1}$, Junjie Hu$^{2}$, Qiuyuan Huang$^{3}$, Lei Zhang$^{3}$, Jana Diesner$^{1}$, Jianfeng Gao$^{3}$ \\
  $^{1}$University of Illinois at Urbana-Champaign, $^{2}$Carnegie Mellon University \\
  $^{3}$Microsoft Research, Redmond\\
  {\tt \{mjiang17,jdiesner\}@illinois.edu, junjieh@cs.cmu.edu} \\
  {\tt \{qihua,leizhang,jfgao\}@microsoft.com}}

\date{}

\begin{document}
\maketitle
\begin{abstract}
  Popular metrics used for evaluating image captioning systems, such as BLEU and CIDEr, provide a single score to gauge the system's overall effectiveness.
  This score is often not informative enough to indicate what specific errors are made by a given system. In this study, we present a fine-grained evaluation method {\metric} for automatically measuring the performance of image captioning systems.
  {\metric} assesses the quality of captions from three perspectives: 1) \textbf{R}elevance to the ground truth, 2) \textbf{E}xtraness of the content that is irrelevant to the ground truth, and 3) \textbf{O}mission of the elements in the images and human references. Experiments on three benchmark datasets demonstrate that our method achieves a higher consistency with human judgments and provides more intuitive evaluation results than alternative metrics.\footnote{Code is released at \url{https://github.com/SeleenaJM/CapEval}.}
\end{abstract}

\section{Introduction}
Image captioning is an interdisciplinary task that aims to automatically generate a text description for a given image.
The task is fundamental to a wide range of applications, including image retrieval \cite{imageretrieval} and vision language navigation \cite{visualnavig}. Though remarkable progress has been made \cite{captiongen1, captiongen2}, the automatic evaluation of image captioning systems remains a challenge, particularly with respect to quantifying the generation errors made by these systems \cite{captionsurvey1}. 

Existing metrics for caption evaluation can be grouped into two categories: 1) rule-based metrics \cite{bleu, cider} that are based on exact string matching, and 2) learning-based metrics \cite{learningmetric1, machine_learn4} that predict the probability of a testing caption as a human-generated caption by using a learning model. In general, prior work has shown that description adequacy with respect to the ground truth data is a main concern for evaluating text generation systems \cite{nlgsurvey}. Though this aspect has been emphasized by prior work for assessing image captions \cite{bleu, meteor, gaosurvey}, one common limitation of existing metrics is the lack of interpretability to the description errors because existing metrics only provide a composite score for the caption quality. Without fine-grained analysis, the developers may not be able to understand the specific description errors made by their developed captioning systems. 

\begin{figure}[t]
    \centering
    \includegraphics[width = 0.48\textwidth]{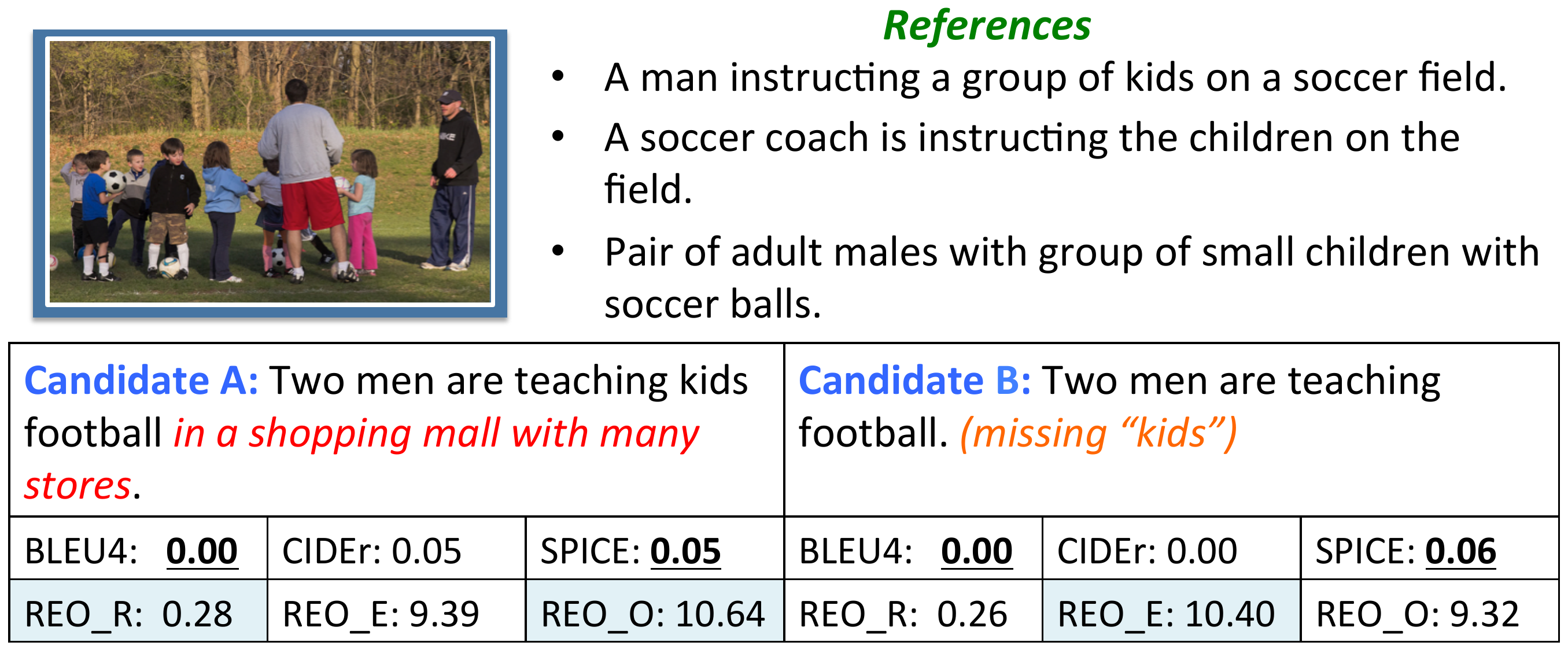}
    \caption{An example of caption evaluation. Given two caption candidates, even though Caption A covers more image information than Caption B (e.g., missing "kids"), Caption A contains extra irrelevant description (highlighted in red). Prior metrics (e.g., BLEU4) only provide an overall quality score, which is difficult to infer specific description mistakes in a caption. In contrast, REO provides three indicators (i.e., relevance, extraness, and omission) that can properly achieve a
    fine-grained assessment for each caption.}
    \label{fig:motivation}
\vspace{-2ex}
\end{figure}

\begin{figure}[t]
    \centering
    \includegraphics[width = 0.47\textwidth]{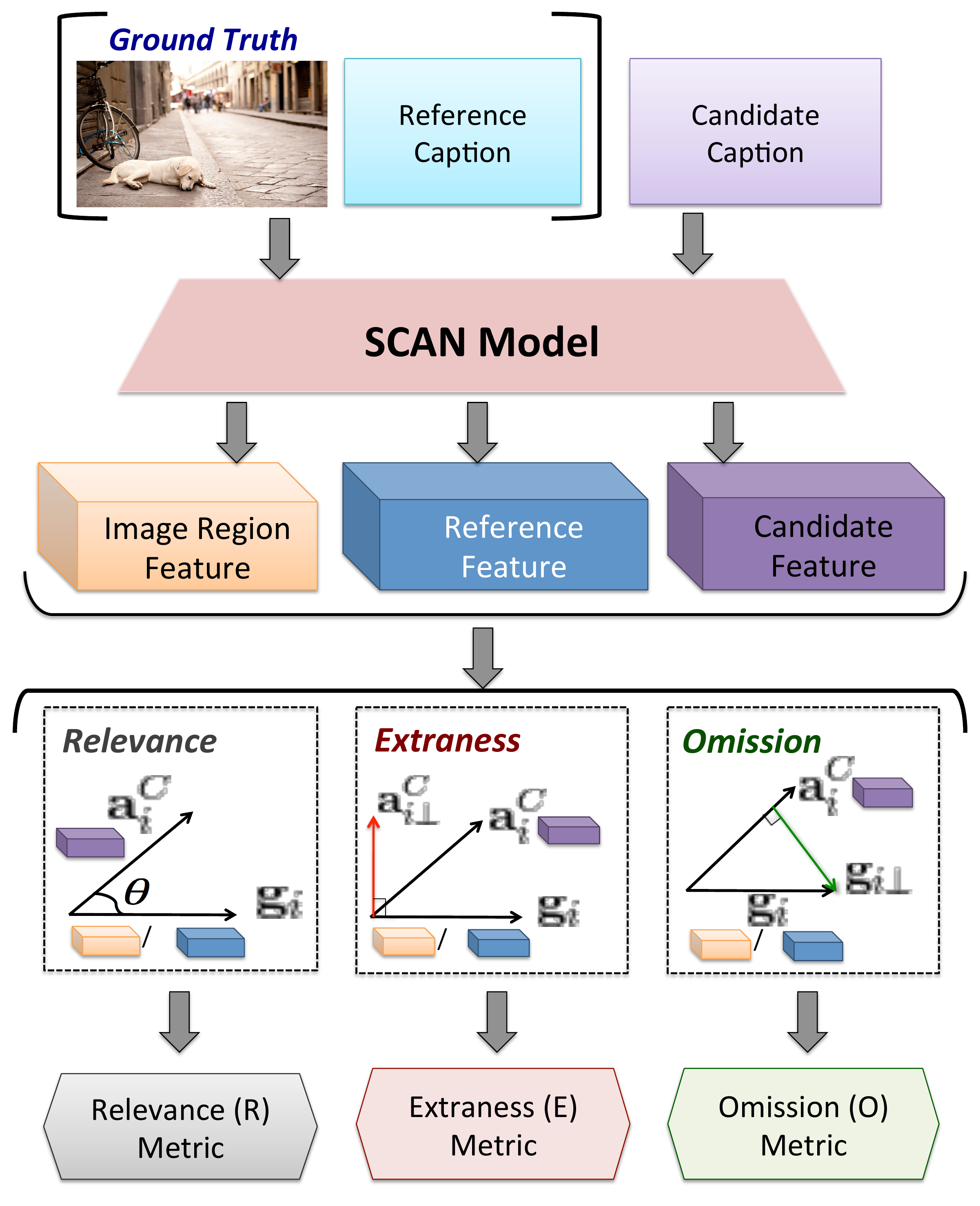}
    \caption{Overview of REO evaluation for image captioning: (1) Feature Extraction; (2) Metric Measurement.}
    \label{fig:method}
\vspace{-2ex}
\end{figure}

To fill this gap, we propose an evaluation method called \textbf{REO} that considers three specific pieces of information for measuring each caption with respect to: 1) \textbf{R}elevance: relevant information of a candidate caption with respect to the ground truth, 2) \textbf{E}xtraness: extra information of a candidate caption beyond ground truth data, and 3) \textbf{O}mission: missing information that a candidate fails to describe from an image and human-generated reference captions. Figure~\ref{fig:motivation} shows a comparison between existing metrics and our proposed metrics that measure caption quality at a fine-grained level.
If we view caption generation as a process of decoding the information embedded in an image, we can evaluate an image captioning system by measuring the effectiveness of the decoding process in terms of the relevance of the decoded information regarding the image content, and the amount of missing or extra information.
Using both the images and reference captions as ground truth information for evaluation, our approach is built based on a shared image-text embedding space defined by a grounding model that has been pre-trained on a large benchmark dataset. Given a pair of vectors representing a candidate caption and its ground truth (i.e., the target image and associated reference captions), respectively, we compute the relevance of the candidate caption and ground truth based on vector similarities. By applying vector orthogonal projection, we identify the extra and missing information carried by the candidate caption. Each aspect that we consider here (i.e., relevance, extraness, omission) is measured by an independent score.

We test our method on three datasets. The experimental results show that our proposed metrics are more consistent with human evaluations than alternative metrics. Interestingly, our study finds that human annotators pay more attention to extra or missing information in a caption (i.e., false positive and false negatives) than the caption's relevance for the given image (true positives). We also find that considering both image and references as ground truth information is more helpful for caption evaluation than considering image or references individually.

\section{Methods}

Figure~\ref{fig:method} provides an overview for the calculation of REO, which happens in two stages. 
The first stage is \textit{feature extraction}, where we aim to obtain feature vectors to encode the candidate caption $C$ and corresponding ground truth $\Gb$ for further comparisons. 
The second stage is to measure \textit{three metric scores}. Specifically, we measure relevance using standard cosine similarity. To measure irrelevance (i.e., extraness and omissions), we compare the information carried by $C$ and $\Gb$, respectively. We will give a detailed description of our method in the following two subsections.

\subsection{Feature Extraction}
Following \citeauthor{scan}, we leverage a pre-trained Stacked Cross Attention Neural Network (SCAN) to build a multi-modal semantic space. Specifically, we obtain word features $\Hb^{\tau} =[\hb_1^{\tau};\cdots; \hb_M^{\tau}]\in \RR^{M\times D}$ by averaging the forward and backward hidden states per word from a bidirectional GRU~\cite{rnngru}, where $\tau=\{C,R\}$ denotes either a candidate $C$ or a reference $R$ sentence of $M$ words. Based on ~\citeauthor{bottomup}, we achieve image features $\Ub \in \RR^{N\times D'}$ by detecting $N$ salient regions per image ($N = 36$ in this paper). A linear layer is applied to transform image features to $D$-dimensional features $\Vb=[\vb_1;\cdots;\vb_N]\in\RR^{N\times D}$.

Based on the SCAN model, we further extract the context information from the caption words for each detected region. To this end, we compute a context feature $\ab_i^{\tau}$ for the $i^{th}$ region by a weighted sum of caption word features in Eq.~(\ref{eq:attend}). Notice that $\Ab^{\tau}=[\ab_1^{\tau};\cdots; \ab_N^{\tau}] \in \RR^{N\times D}$ extracts the context information of the caption $\tau$ with respect to all regions in the image. 
\vspace{-2.0ex}
\begin{align}
\label{eq:attend}
\mathbf{a}^{\tau}_i &= \sum_{j=1}^m \alpha_{ij} \mathbf{h}^{\tau}_j \\
\quad \alpha_{ij} &= \frac{\exp(\lambda \text{sim}(\mathbf{v}_i, \mathbf{h}^{\tau}_j))}{\sum_{k = 1}^m \exp(\lambda \text{sim}(\mathbf{v}_i, \mathbf{h}^{\tau}_k))}
\end{align}
where $\lambda$ is a smoothing factor, and $\text{sim}(\mathbf{v}, \mathbf{h}^{\tau})$ is a normalized similarity function defined as
\vspace{-1ex}
\begin{equation}\label{eq2}
    \text{sim}(\mathbf{v}_i, \mathbf{h}^{\tau}_j) = \frac{\max (0, \text{cosine}(\mathbf{v}_i, \mathbf{h}^{\tau}_j))} {\sqrt{\sum_{k = 1}^n \max(0, \text{cosine}(\mathbf{v}_k, \mathbf{h}^{\tau}_j)) ^2}}
\end{equation}
\vspace{-4ex}
\subsection{Metric Scores}
In order to explore the impact of image data on evaluation, we focus on comparing context features $\Ab^C$ of the candidate caption $C$ to ground-truth references $\Gb \equiv [\gb_1;\cdots;\gb_N] = \cbr{\Vb, \Ab^{R}}$, where $\Gb$ denotes either image features $\Vb$ or the context features of $R$ (i.e., $\Ab^R$).

\paragraph{Relevance}: The relevance between a candidate caption and a ground-truth reference based on the $i$-th region is computed by the cosine similarity of $\ab^C_i$ and $\gb_i$. We average similarity over all regions to get the relevance score of a candidate caption with respect to an image. 
\vspace{-1ex}
\begin{align}
    \label{eq:relevance}
    \Rcal = {1\over N}\sum_{i=1}^N \text{sim}(\mathbf{a}_i^C, \mathbf{g}_i)   
\end{align}

\paragraph{Extraness}: The extraness of $C$ is captured by performing an orthogonal projection of $\ab_i^C$ to $\gb_i$, which returns the vertical context vector $\ab_{i\perp}^C$ to represent the irrelevant content of $C$ to the ground truth at the $i^{th}$ region.
\vspace{-1ex}
\begin{align}
    \mathbf{a}_{i\perp}^C = \mathbf{a}_i^C - \frac{\mathbf{a}_i^C \cdot \mathbf{g}_i}{\parallel \mathbf{g}_i \parallel^2}\mathbf{g}_i .
\end{align}
To avoid potential disturbance due to correlated feature vectors, we measure the Mahalanobis distance between the vertical context vector $\ab_{i\perp}^C$ and its original context vector $\ab_i^C$ (see Eq.~(\ref{eq:dist})). Notice that a small distance value indicates that the irrelevant context vector $\ab_{i\perp}^{C}$ is closed to the original context vector $\ab_{i}^{C}$. In other words, the original context contains a large amount of extra information. Therefore, the higher this metric is, the less extra information the caption contains. 
\vspace{-1ex}
\begin{align}
\Ecal = {1\over N} \sum_{i=1}^N d( \ab_i^C, \ab_{i\perp}^C) 
\end{align}
\vspace{-2.5ex}
\begin{align} \label{eq:dist}
d(\mathbf{p}, \mathbf{q}) = \sqrt{(\mathbf{p}-\mathbf{q})S^{-1}(\mathbf{p}-\mathbf{q})} 
\end{align}

\paragraph{Omission}: The measurement of omission is similar to that of extraness, where we capture the missing information of $C$ by the vertical context features $\gb_{i\perp}$ based on the orthogonal projection of $\gb_i$ to $\ab^C_i$. The omission score is denoted as $\Ocal$. Similarly, the higher the omission score is, the less missing information the caption contains.
\vspace{-1.5ex}
\begin{align}
\mathbf{g}_{i\perp} &= \mathbf{g}_i - \frac{\mathbf{g}_i \cdot \mathbf{a}_i^C}{\parallel \mathbf{a}_i^C \parallel^2}\mathbf{a}_i^C \\
    \Ocal &= {1\over N} \sum_{i=1}^N d(\gb_i, \gb_{i\perp})
\end{align}

Considering that each image may have multiple reference captions, we further average the score of the aforementioned three aspects over all reference captions while considering $\Ab^R$ as ground truth.

\begin{table*}[t]
\centering
\resizebox{2.0\columnwidth}{!}{%
\begin{tabular}{ l l  c  c||c  c  c  c  c}
    \toprule
    \multirow{2}{*}{\textbf{Ground Truth}} & \multirow{2}{*}{\textbf{Metric}} & \textbf{Composite } & \textbf{PublicSys} & \multicolumn{5}{c}{\textbf{Pascal-50S $(accuracy\%)$}}\\
    & & $(\tau)$ & $(\tau)$ & \textbf{HC} & \textbf{HI} & \textbf{HM} & \textbf{MM} & \textbf{ALL} \\ 
	\hline
	\hline
	\multirow{6}{*}{Reference} & BLEU-1  &  0.280 & 0.267 & 50.50          & 94.50          & 92.30          & 56.00          & 73.33          \\ 
	&BLEU-4  &  0.205 & 0.223  & 50.60          & 91.90          & 85.60          & 60.90          & 72.25          \\
	&ROUGE\_L & 0.307 & 0.232 & 53.30          & 94.60          & 93.50          & 58,20          & 74.90          \\
	&METEOR &  0.379 & 0.254   & 58.00          & 97.60          & 94.90          & 63.40          & 78.48          \\
	&CIDEr  &  0.378 & 0.278  & 54.80          & 97.90          & 91.50          & 63.80          & 77.00          \\
	&SPICE  &  0.419 & 0.258   & 56.60          & 94.70          & 85.00          & 49.00          & 71.33          \\
    \midrule
    \multirow{3}{*}{Image (ours)} & Relevance  & 0.423 & 0.148  & 58.40          & 99.40          & 93.10          & 73.40          & 81.08          \\ 
    &Extraness  & 0.430 & 0.149   & 56.80          & \textbf{99.70} & 92.80          & 74.80          & 81.03          \\
    &Omission  &  0.445 & 0.165   & \textbf{61.00} & 99.40          & 93.80          & 69.10          & 80.83          \\
    \midrule
    \multirow{3}{*}{\makecell{Image + Reference \\ (ours)}} & Relevance &  0.502 & 0.313  & 56.40          & \textbf{99.70} & 93.50          & 77.10          & 81.68          \\ 
    &Extraness &  0.507 & \textbf{0.320}  & 54.30          & 99.60          & 92.60          & \textbf{77.20} & 80.93          \\
    &Omission & \textbf{0.533} & 0.291 & 60.00          & 99.60          & \textbf{95.40} & 72.50          & \textbf{81.88} \\
    \bottomrule
\end{tabular}
}
\caption{Caption-level correlation between metrics and human grading scores in Composite and PublicSys dataset by using Kendall tau $(\tau)$. All p-values \(<\) 0.01. For PASCAL-50S, we display the accuracy of metrics at matching human judgments with 5 reference captions per image. The highest value per column is in bold font. Column titles are explained in Section 3.1. Ground truth refers to two points of information: human-written references and images.}
\label{tab:result}
\end{table*}

\section{Experiments}
\subsection{Experimental Setup}
We perform experiments on three human-evaluated caption sets. \textit{\textbf{Composite Dataset}} \cite{composite} contains the candidate captions of images from MS-COCO, Flickr8k, and Flickr30k. Captions were generated by humans and two caption models (11,985 instances in total). Human judgments for these candidate captions was provided on a 5-point scale rating that represents description correctness. \textit{\textbf{PublicSys Dataset}} has 2,500 captions collected by \citeauthor{Hallucination}, which were generated by five state-of-the-art captioning systems on 500 MS-COCO images, respectively. Human grading was done on a 5-point scale based on annotators' preferences to descriptions. \textit{\textbf{Pascal-50S Dataset}} \cite{cider} includes 4000 caption pairs that describe images from the UIUC PASCAL Sentence dataset. Each annotator was asked to select one sentence per pair that is closer to the expression of the given reference sentence. Candidate pairs were grouped into four categories: 1) human-human correct  (HC, i.e., a pair of captions are written by humans for the same image), 2) human-human  incorrect  (HI, i.e., Two human-written captions of which one describes another image instead of the target image), 3) human-machine  (HM, i.e., two captions are generated by a human and a machine, respectively.),  and 4)  machine-machine  (MM, i.e., both machine-generated caption).

Following standard practice \cite{spice, evlauation1}, we compared with rule-based metrics (see Table~\ref{tab:result}). All existing metrics were implemented with the MS-COCO evaluation tool \footnote{https://github.com/tylin/coco-caption}. The performance of metrics was assessed via Kendall's tau $(\tau)$ rank correlation for the scoring-based datasets (i.e., Composite \& PublicSys) and accuracy of the pairwise comparison for the Pascal-50S dataset.
\begin{figure}
    \centering
    \includegraphics[width = 0.49\textwidth]{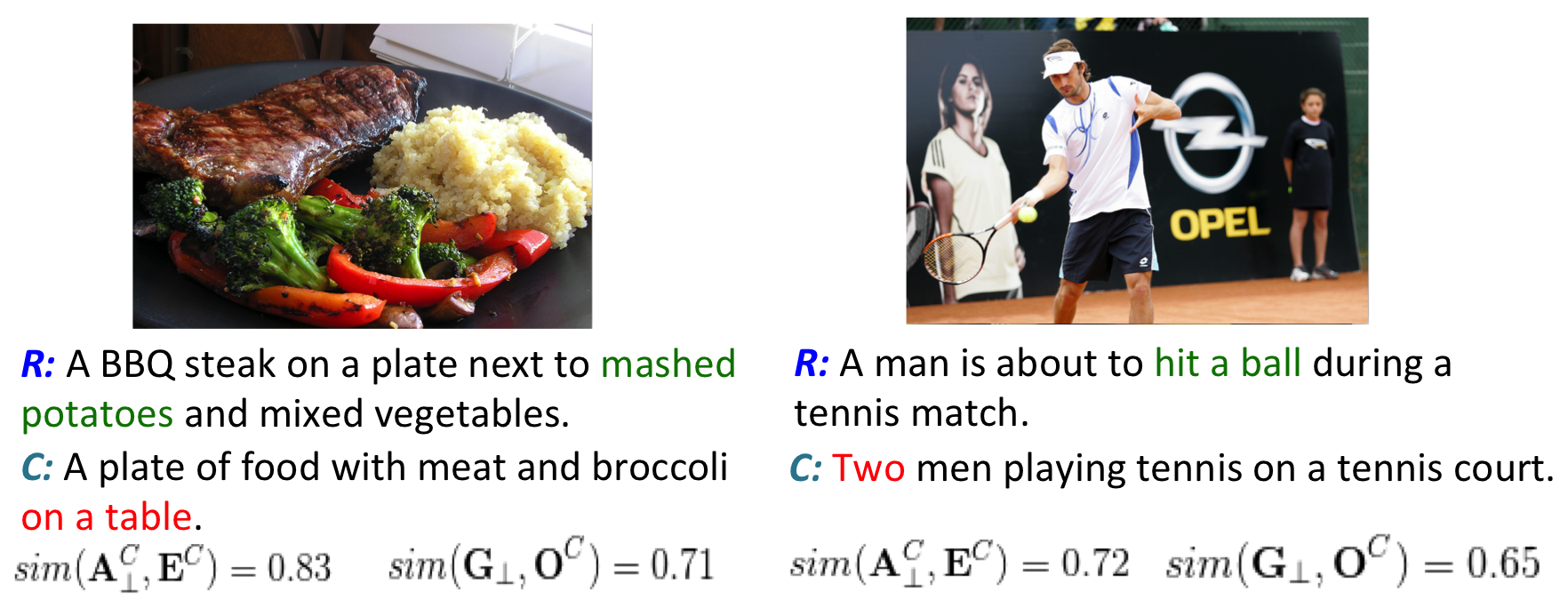}
    \vspace{-4ex}
    \caption{Examples of validating error identification. Text in red is extra information, while text in green is missing information. $sim(x,y)$ is the average similarity between machine-identified and true error vectors over image regions.}
    \label{fig:fpfn}
\end{figure}

\begin{figure*}[t]
    \centering
    \includegraphics[width = \textwidth]{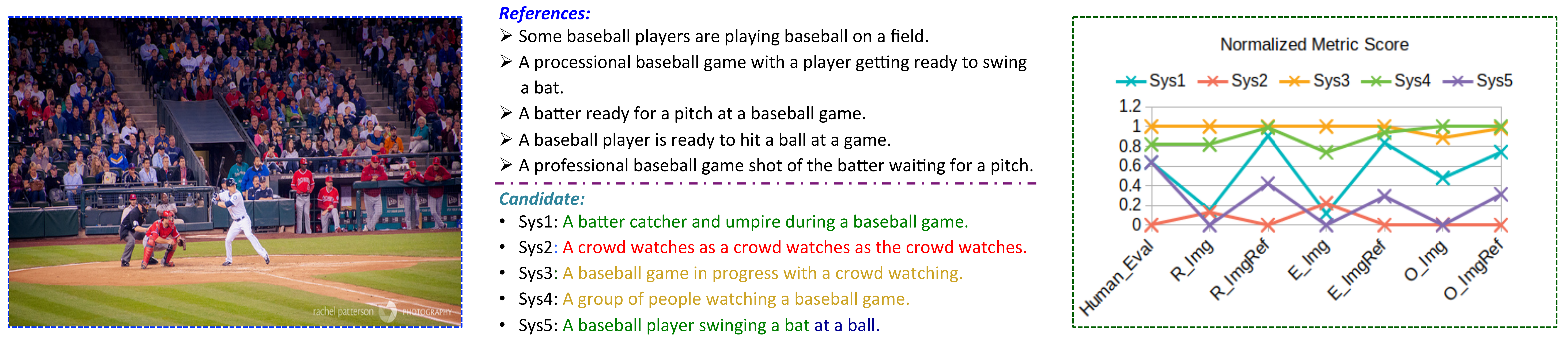}
    \vspace{-1 ex}
    \caption{Case study of REO metric scores. Candidates are highlighted in: 1) green: a detailed but incompleted caption, 2) red: repetition, 3) yellow: high-level description, and 4) blue: extra information not shown in the image.}
    \label{fig:case}
\end{figure*}

\subsection{Experimental Results}
\paragraph{Can extra \& missing information be captured?}
In order to measure the effectiveness of error identification (i.e., extraness and omission), we randomly sampled a subset of data, and manually identify the actual extraness (i.e., $\Eb^C$) and true omission (i.e., $\Ob^C$) of each candidate caption. We conduct validation based on the average cosine similarity between the machine-identified error (i.e., $\ab_{i\perp}^C \& \gb_{i\perp}$) and true error description.

Figure~\ref{fig:fpfn} provides two illustrative examples of the validation process. Phrases highlighted in red (e.g.,"on a table") are extra information (more text in the candidate caption than in the ground truth). Meanwhile, phrases in green (e.g., "mashed potatoes") are missing from the candidate description, but occur in the image and the reference caption.  We observe that machine-identified errors are highly similar to the true error information in both cases ($\geq 0.65$). This result suggests that our method can capture extraness and omission from an image caption.

\paragraph{Do error-aware evaluation metrics help?} The results of metric performance in Table~\ref{tab:result} show that overall, using the three metrics proposed in REO, especially extraness and omission, led to a noticeable improvement in Kendall tau's correlation compared to the best reported results based on prior metrics. Our results suggest that human evaluation tends to be more sensitive to the irrelevance than the relevance of a candidate caption regarding ground truth. We also find that jointly considering both images and human-written references contributes more to caption evaluation than each of the two data sources alone - except for the case of HC pair comparison. This exception can be explained by the phenomenon that human-written descriptions are flexible in terms of word choice and sentence structure, and such diversity may lead to the challenge of comparing a reference to a candidate in cases where both captions were provided by humans. By further looking into each considered aspect in REO, we find that extranesss metric is more appropriate to evaluate machine-generated captions (e.g., the PublicSys dataset and the MM pairs of Pascal-50S dataset), while the omission metric can be a better choice to access caption quality when the testing data consists of human-written descriptions.

\paragraph{What can we learn from the metric outputs?} To analyze our metric scores more in depth, we compare the outputs of five captioning systems on a set of images in the PublicSys dataset. Figure~\ref{fig:case} shows an illustrative example. To make the scale of human grading and REO metrics comparable, we normalize scores per metric by using max-min normalization. 

We find that metrics calculated in cases where the ground truth contained both the target image and human references are more likely to identify expression errors. For example, though the phrase ``a crowd watches'' in the caption of system 2 is relevant to the image, this phrase is repeated by three times in a sentence. As a result, the scores for relevance and extraness are decreasing when the ground truth involves references. Also, metrics focusing only on image content return higher values when the testing captions provide a high-level description of the whole image (e.g., captions of system 3 and 4) compared to the detailed captions for a specific image part (e.g., captions of system 1 and 5 focus on the baseball player). By comparing the herein considered three aspects of each caption, we observe that a caption that mainly focuses on describing a part of image in detail boosts relevance, but the sentence achieves a reduced metric score in terms of omission. 


\section{Conclusion}
This paper presents a fine-grained, error-aware evaluation method {\metric} to measure the quality of machine-generated image captions according to three aspects of descriptions: relevance regarding ground truth, extra description beyond image content, and omitted ground truth information. Comparing these metrics to alternative metrics, we find that our proposed solution produces evaluations that are more consistent with the assessment of human judges. Moreover, we find that human judgment tends to penalize extra and missing information (false positives and false negatives) more than it appreciates relevant content. Finally, and to no surprise, we conclude that using a combination of image content and human-written references as ground truth data allows for a more comprehensive evaluation than using either type of information separately. 
Our method can be extended to evaluate other text generation tasks.

\section*{Acknowledgments}
We appreciate anonymous reviewers for their constructive comments and insightful suggestions. This work was partly performed when Ming Jiang was interning at Microsoft Research. The authors would like to thank Pengchuan Zhang for his help with pre-training the grounding model.

\bibliography{reo}
\bibliographystyle{acl_natbib}

\appendix


\end{document}